\documentclass{bmvc2k}

\usepackage{graphicx}
\usepackage{amsmath}
\usepackage{array}
\usepackage{amssymb}
\usepackage{placeins}
\usepackage{xcolor,soul}
\usepackage[framemethod=tikz]{mdframed}
\usepackage[ruled,vlined]{algorithm2e}
\usepackage{multirow} % Merge table cells vertically
\usepackage{caption} % Fancy captions for tables
\usepackage{wasysym}
\usepackage{booktabs} % Fancy table rules
\usepackage[activate={true,nocompatibility},final,tracking=true,kerning=true,spacing=true,factor=1100,stretch=10,shrink=9]{microtype}
\usepackage[accsupp]{axessibility}  % Improves PDF readability for those with disabilities.
\usepackage{subcaption}
\usepackage{enumitem}% http://ctan.org/pkg/enumitem
\usepackage[toc,page]{appendix}

\newcommand{\squeezeupsmall}{\vspace{-1mm}}
\newcolumntype{C}{>{\centering\arraybackslash}p{0.06\textwidth}}

\title{ZeST-NeRF: Using temporal aggregation for Zero-Shot Temporal NeRFs}

\addauthor{Violeta Menéndez González}{v.menendezgonzalez@surrey.ac.uk}{1,2}
\addauthor{Andrew Gilbert}{a.gilbert@surrey.ac.uk}{1}
\addauthor{Graeme Phillipson}{graeme.phillipson@bbc.co.uk}{2}
\addauthor{Stephen Jolly}{stephen.jolly@bbc.co.uk}{2}
\addauthor{Simon Hadfield}{s.hadfield@surrey.ac.uk}{1}
\addinstitution{
 Centre for Vision, Speech and Signal Processing (CVSSP)\\
 University of Surrey\\
 Guildford, UK
}
\addinstitution{
 BBC R\&D\\
 MediaCityUK\\
 Salford, UK
}

\runninghead{Menéndez González \etal}{ZeST-NeRF}

\def\eg{\emph{e.g}\bmvaOneDot}

\def\etal{\emph{et al}\bmvaOneDot}

\begin{document}

\maketitle

\begin{figure}[htbp]
        \vspace{-0.8cm}
	\centering
	\begin{subfigure}[b]{0.45\textwidth}
  		\centering
  		\includegraphics[width=\textwidth]{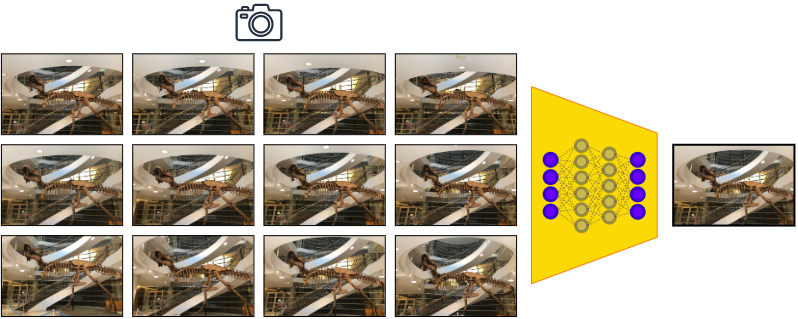}
  		\caption{Scene-specific image (\eg NeRF~\cite{mildenhall2020nerf})}
        \label{subfig:teaser_NeRF}
	\end{subfigure}
	\begin{subfigure}[b]{0.54\textwidth}
  		\centering
  		\includegraphics[width=\textwidth]{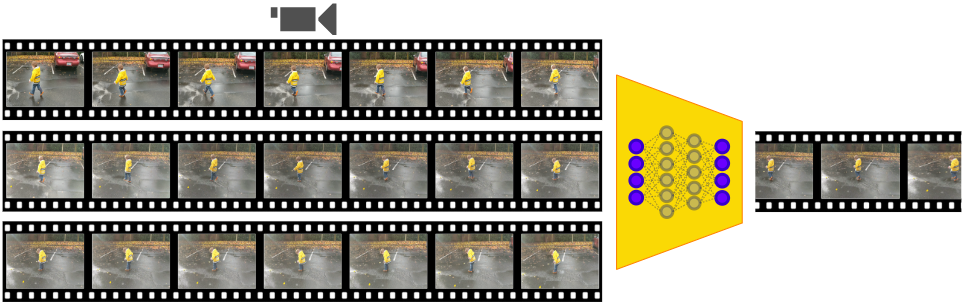}
  		\caption{Scene-specific video (\eg NSFF~\cite{li2021NSFF})}
        \label{subfig:teaser_NSFF}
	\end{subfigure}
	\begin{subfigure}[b]{0.45\textwidth}
  		\centering
  		\includegraphics[width=\textwidth]{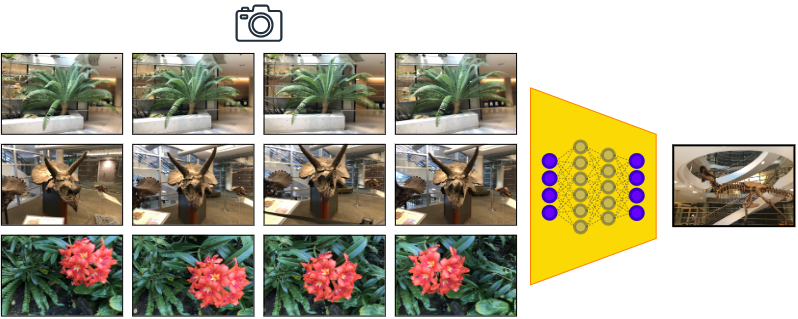}
  		\caption{``Zero-shot'' image (\eg MVSNeRF~\cite{chen2021mvsnerf})}
        \label{subfig:teaser_MVSNeRF}
	\end{subfigure}
	\begin{subfigure}[b]{0.54\textwidth}
  		\centering
  		\includegraphics[width=\textwidth]{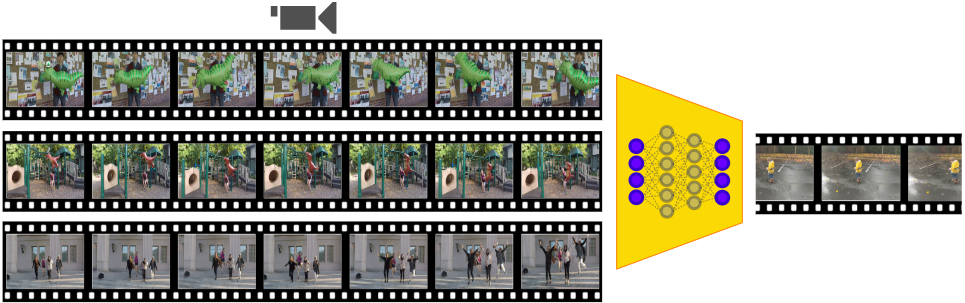}
  		\caption{``Zero-shot'' video (ZeST-NeRF)}
        \label{subfig:teaser_ZeST}
	\end{subfigure}
        \vspace{0.01cm}
	\caption{\textbf{``Zero-shot'' temporal NeRF} contrasted with existing techniques}
	\label{fig:teaser}
\end{figure}
\vspace{-0.7cm}
\begin{abstract}
In the field of media production, video editing techniques play a pivotal role. Recent approaches have had great success at performing novel view image synthesis of static scenes. But adding temporal information adds an extra layer of complexity. Previous models have focused on implicitly representing static and dynamic scenes using NeRF. These models achieve impressive results but are costly at training and inference time. They overfit an MLP to describe the scene implicitly as a function of position. This paper proposes ZeST-NeRF, a new approach that can produce temporal NeRFs for new scenes without retraining.
We can accurately reconstruct novel views using multi-view synthesis techniques and scene flow-field estimation, trained only with unrelated scenes. We demonstrate how existing state-of-the-art approaches from a range of fields cannot adequately solve this new task and demonstrate the efficacy of our solution. The resulting network improves quantitatively by 15\% and produces significantly better visual results.
\end{abstract}
%-------------------------------------------------------------------------
\section{Introduction}
\label{sec:intro}
Producing photo-realistic renderings of natural scenes under new viewpoints is important for media content production, virtual/augmented reality, and image/video editing. The challenge increases significantly when the scene is time-dependent, i.e. when the camera or subjects undergo movement. 

In the area of novel view synthesis, many new approaches~\cite{niemeyer2022regnerf,barron2022mipnerf360} rely on the popular implicit representation given by Neural Radiance Fields (NeRF)~\cite{mildenhall2020nerf} (Fig.~\ref{subfig:teaser_NeRF}). This has been extended to dynamic scene methods that aim to model the temporal dimension alongside the spatial ones~\cite{li2021NSFF,li2022DyNeRF} (Fig.~\ref{subfig:teaser_NSFF}). These approaches achieve impressive photo-realistic results but have similar limitations to NeRF, being very expensive, requiring a lot of input views, and having a very long per-scene optimisation process. Some approaches~\cite{yu2021pixelnerf,chen2021mvsnerf,menendez2022svs} worked towards generalising NeRFs to unseen scenes and reducing the required number of input views (Fig.~\ref{subfig:teaser_MVSNeRF}). This is already a highly ill-defined problem, which suffers from uncertainty and ambiguity in the reconstructed data. Naively adding a layer of temporal complexity can significantly exacerbate this. In this work, the temporal data becomes an advantage, as information is aggregated across frames to resolve ambiguities in unknown scenes.

In particular, we propose ZeST-NeRF (Fig.~\ref{subfig:teaser_ZeST}), which uses temporal information to reconstruct a scene geometry and motion estimate. This can then be used to generalise to unseen scenes without requiring expensive retraining. Given a set of keyframes, we use a geometry encoding volume to inform a static NeRF (which reconstructs the background) and a dynamic encoding volume to inform a dynamic NeRF (which reconstructs the motion between frames). We then combine these to generate a new video from a new point of view. Because our approach learns to estimate structure and motion in a scene-agnostic way, it can be applied ``zero-shot''\footnote{Note that in the context of NeRFs we use 'zero-shot' to refer to the ability to perform inference on scenes (and scene configurations) that differ from those seen during training. This is in slight contrast to the usage of the phrase in supervised learning which refers specifically to 'categories' of data which were unseen during training} to new scenes without laborious training. It uses temporal and spatial information to recover missing information in unseen areas. It can be further fine-tuned for short amounts of time to increase the quality of the reconstructions. We will make our code publicly available\footnote{https://github.com/violetamenendez/zest-nerf}.
In summary, the contributions of this paper are:
\begin{itemize}[itemsep=1pt,topsep=4pt]
    \item First radiance field approach capable of ``zero-shot'' novel viewpoint rendering in completely unseen videos of complex scenes.
    \item An efficient multi-encoding-volume approach to scene-agnostic video representation.
    \item A new evaluation protocol and newly developed baselines for the problem of ``zero-shot'' novel-viewpoint-rendering in the video.
\end{itemize}

%-------------------------------------------------------------------------
\section{Background}
\paragraph{Novel view synthesis}
Classic Image-based rendering (IBR) techniques \cite{mcmillan1995,debevec1998,chaurasia2013} usually attempt to model an intermediate scene geometry. These geometrical representations are based on restrictive structures like voxel grids~\cite{sitzmann2019deepvoxels,shi2021}, point clouds~\cite{wiles2020synsin, xu2022pointNeRF}, or multi-layer arrangements~\cite{flynn2016deepstereo,xu2019, zhou2018stereomag,flynn2019deepview}. In recent years, implicit representations have been prevalent when approaching novel view synthesis techniques. NeRF~\cite{mildenhall2020nerf} proposed an entirely neural scene representation. A Multi-Layer Perceptron (MLP) is used to parameterise a function rendering density and colour by querying 3D location and viewing direction. This approach revolutionised the field; however, in its original form, NeRF is very costly to run. In addition, this model has a long per-scene optimisation process, which prevents it from being useful in many important applications. Following approaches worked towards improving performance and increasing flexibility~\cite{martin-brualla2021,barron2021mipnerf,srinivasan2021, hu2022}. Regardless, these models require dense input views, are costly, and don't generalise to new scenes.

Some recent approaches have relaxed the requirement for a high number of input views by applying data augmentation~\cite{chen2022a} or by introducing regularisations~\cite{kim2022, rebain2022lolnerf, niemeyer2022regnerf, deng2022}. Furthermore, some approaches attempt to generalise their models to apply to new scenes that the models haven't been trained on~\cite{yu2021pixelnerf, johari2022geonerf}. This is achieved by introducing neural geometry priors based on IBR approaches, like 2D feature aggregation~\cite{wang2021ibrnet}, stereo-matching techniques~\cite{chibane2021srf}, or recurrent aggregation~\cite{zhang2022nerfusion}. MVSNeRF~\cite{chen2021mvsnerf} constructs a 3D feature encoding volume based on Plane Sweep Volumes~\cite{flynn2016deepstereo}. The model only needs three input images at training time and can still generalise to new scenes. Nevertheless, these models suffer from significant artefacts and incorrect outputs in occluded areas that are not visible from the reference view. SVS~\cite{menendez2022svs} tries to solve this by applying adversarial training to hallucinate content in the regions occluded in all inputs. While this improves results in many cases, the adversarial regime harms training stability. Our paper extends this `encoding volume' approach to ``zero-shot'' transfer, developing both a static and dynamic encoding volume for application to temporal sequences.

\paragraph{Dynamic scene representation}
Within dynamic scene reconstruction, many approaches require ground truth RGBD or costly hand-labelled data.
Some methods concentrate on recovering either a single non-rigid 3D object~\cite{bozic2020DeepDeform,dou2016fusion4D,innmann2016VolumeDeform,newcombe2015dynamicfusion,zollhofer2014, huang2018, pumarola2021DNeRF, zhao2022HumanNeRF,kwon2021NHP}, or sparse geometry and motion~\cite{park2010,simon2017,vo2016,zheng2015}. Yet others leverage monocular depth estimation models and image segmentation approaches to decompose scenes into rigid and non-rigid areas~\cite{kumar2017SuperPixelSoup,ranftl2016dyMidas,russell2014, luo2020a, xian2021}. Some transform a single canonical radiance field with rigid transformations or modelled dynamics~\cite{du2021,park2021nerfies,pumarola2021DNeRF,tretschk2021,yuan2021STaR,park2021HyperNeRF}. And others attempt to use other pre-computed scene representations to improve rendering efficiency~\cite{lin2021Deep3DMaskVol,xing2022tempMPI}.
Recently some techniques have explored estimating scene flow \cite{hadfield2013} for this \cite{hur2020,brickwedde2019MonoSF,jiang2019SENSE,lv2018,mittal2020,sun2015}. 

When reconstructing scene geometry for novel view synthesis, many approaches require multiple input views from synchronised videos~\cite{zitnick2004,bansal2020,bemana2020xfield}. Yoon~\etal~\cite{yoon2020DFNet} suggest a self-supervised depth fusion network DFNet. This model combines dense view-dependent monocular depth and sparse view-independent multi-view stereo to explain dynamic scene geometry. However, this approach required synchronised annotated data and produced significant artefacts in disocclusions. Instead, other approaches~\cite{li2021NSFF, gao2021} propose a space-time synthesis model that extends NeRF~\cite{mildenhall2020nerf} to estimate scene flow fields. They deal with occlusions by predicting occlusion weights or having 3D flow supervision. Still, they suffer from similar computational limitations as other NeRF models. Li~\etal~\cite{li2022DyNeRF} use importance sampling and hierarchical training to boost training speed. Nevertheless, all these methods require per-scene optimisation processes. In contrast, we aim to render new scenes the model hasn't trained on in a ``zero-shot'' manner.
\squeezeupsmall
%-------------------------------------------------------------------------
\section{Approach}
\begin{figure*}[htbp]
    \begin{center}
        \includegraphics[width=\linewidth]{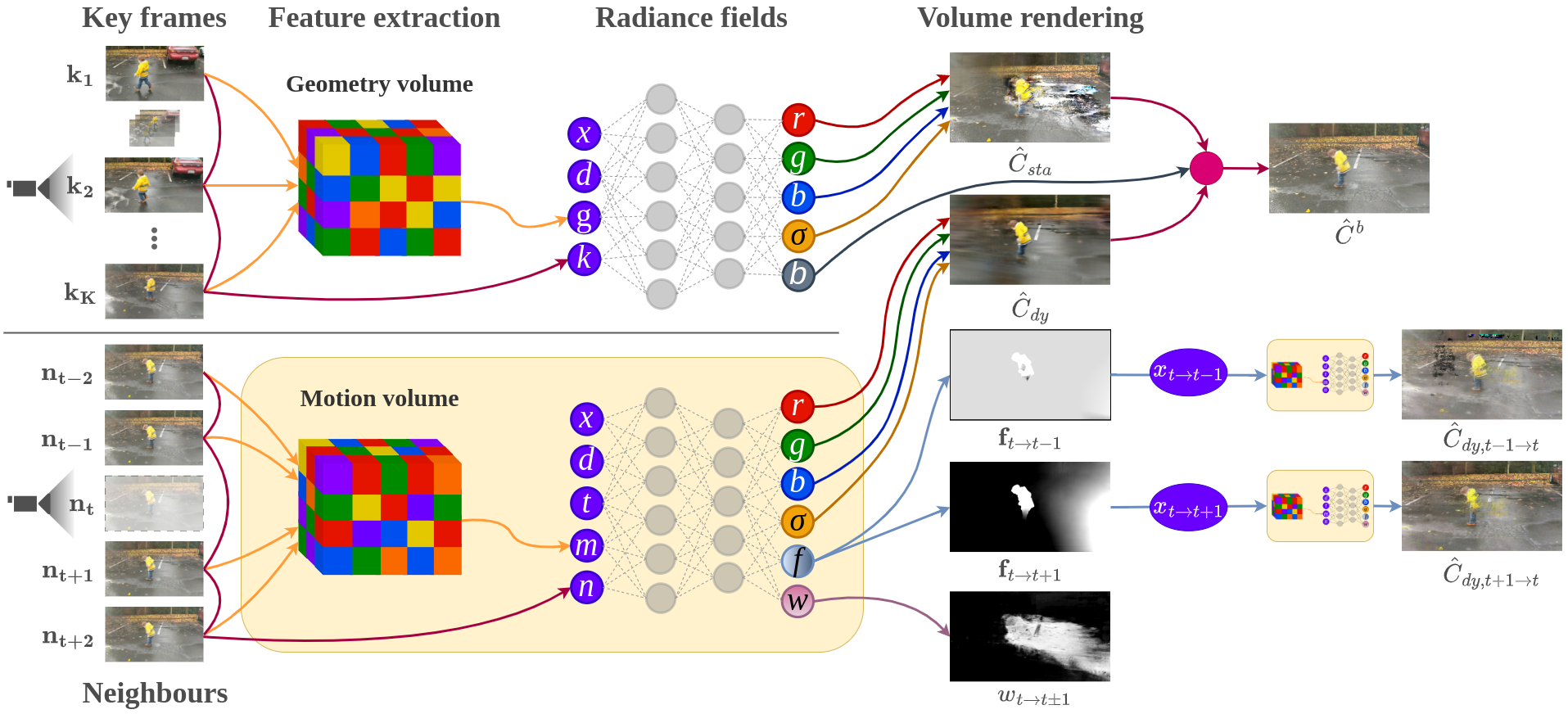}
    \end{center}
    \caption{\textbf{Model overview:} ZeST-NeRF uses a geometry and a motion encoding volume to inform the static and the dynamic NeRFs, respectively.}
   \label{fig:overview}
\end{figure*}

\subsection{Geometry and motion encoding volumes}
Our architecture (See Fig.~\ref{fig:overview}) exploits the power of multi-view 3D CNN encoding volumes~\cite{chen2021mvsnerf, menendez2022svs} to generalise to new scenes. This 3D volume integrates 2D CNN features of the input images by warping multiple sweeping planes of source view features. This differs from techniques like Deep Stereo~\cite{flynn2016deepstereo}, which perform plane sweeps using the raw colour pixels to produce their correlation volume. This allows us to generalise correlations between images, which can then be used to reason about geometry and motion, helping the network generalise to previously unseen scenes.

We use two different volumes which learn different correlations between frames by construction. The \emph{geometry volume} and the \emph{motion volume} described below are constructed with the same network architecture but have different inputs and do not share weights.

\paragraph{Geometry volume}
Given the image sequence $\mathcal{V}=\{\mathbf{I}_{i}\mid \mathbf{I}_{i} \in \mathbb{R}^{H \times W \times 3}\}_{i=1}^{N}$, we use \emph{K} key-frames that are representative of the whole video $\mathcal{K}=\{\mathbf{I}_{i}\mid \mathbf{I}_{i} \in \mathcal{V}, i=\lambda \cdot j, \lambda \in \mathbb{N}\}_{j=1}^{K}$. In our experiments, we use key-frames equally spread across the sequence. However, our approach could easily make use of intelligent key-frame selection techniques. These frames will inform the volume about the general geometry of the scene, particularly including static background elements. We then extract the deep features of each key frame using a deep 2D convolutional network $E(\mathbf{I}_i | \mathbf{w}_\Psi)$ with weights $\mathbf{w}_\Psi$. This network consists of downsampling convolutional layers, batch-normalisation and ReLU activation layers.

We build the plane sweep volume by aligning each feature map to the reference view at multiple depths. To achieve this, a homography $\mathcal{H}_i\left(d\right)$ is computed for each view at each depth.
Given the camera parameters $\{\mathbf{K}_i,\mathbf{R}_i,\mathbf{t}_i\}$ (intrinsics, rotation and translation) for camera $i$ the homography is defined as 
\begin{equation} \label{eq:homography}
    \mathcal{H}_i\left(d\right)=\mathbf{K}_i \cdot \mathbf{R}_i \cdot \left(\mathbb{I}+\frac{\left(\mathbf{t}_{ref}-\mathbf{t}_i\right)\cdot\mathbf{n}_{ref}^{T}}{d}\right)\cdot\mathbf{R}_{ref}^{T}\cdot \mathbf{K}_{ref}^{T}
\end{equation}
where $\mathbb{I}$ is the $3\times3$ identity matrix, $\mathbf{n}_{ref}$ the principle axis of the reference camera, and $d$ is the depth to which the images are being warped. This operation is differentiable, which allows for end-to-end training of the feature encoding network weights $\mathbf{w}_\Psi$ based on the downstream reconstruction losses.

The feature sweep volumes $S(\mathbf{I}_{i})=\{W\!\left(\mathcal{H}_i\left(d\right), E(\mathbf{I}_i | \mathbf{w}_\Psi)\right) \mid \forall d \in \left\{1,...,D\right\}\}$ for each keyframe $\mathbf{I}_i \in \mathcal{K}$ are created by applying the warping function $W$ determined by the homography $\mathcal{H}$ to the keyframe's feature maps at every depth.
Then, a variance-based cost volume $\mathbf{C}(\mathcal{K})=Var\left(\{S(\mathbf{I}_{i})| \forall \mathbf{I}_{i} \in \mathcal{K}\}\right)$~\cite{cheng2020deepstereo,yao2018mvsnet} is generated by aggregating all the warped feature sweep volumes.
This cost volume encodes appearance variations across views. A variance based metric makes it possible to compute this using an arbitrary number of input keyframes.

\begin{figure*}[htbp]
  \centering
   \includegraphics[width=1\linewidth]{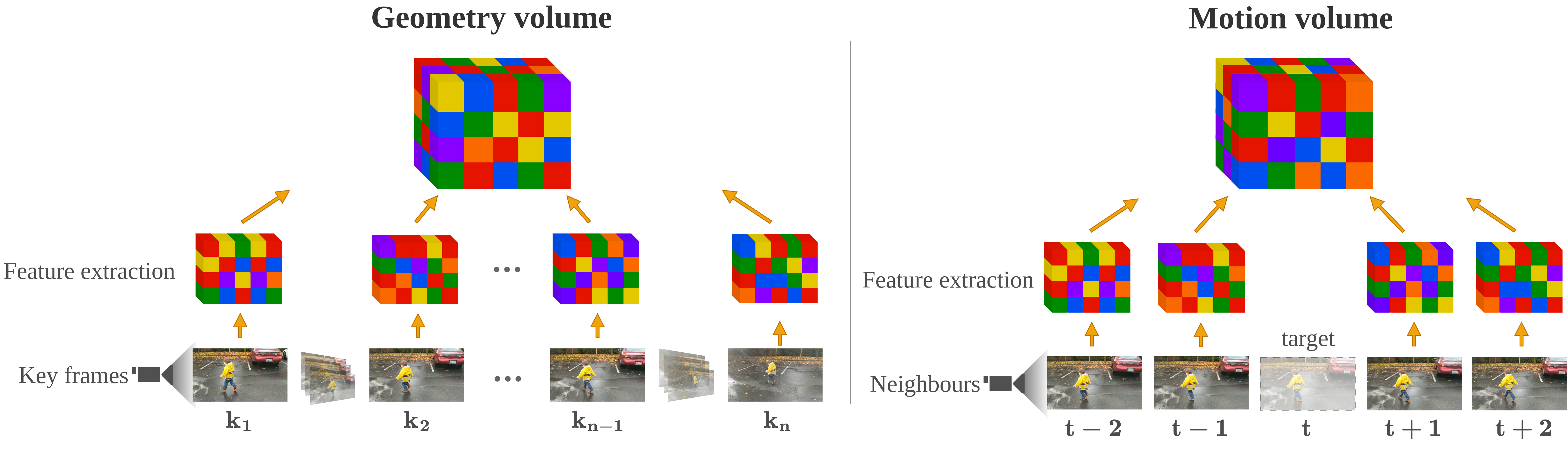}

   \caption{\textbf{Volumes:} Static (left) and Dynamic (right) encoding volumes}
   \label{fig:volumes}
\end{figure*}

Finally, the cost volume is processed using a 3D CNN UNet-like network~\cite{ronneberger2015unet}. The output of this network is the neural geometry volume $\mathbf{G}=V(\mathbf{C}(\mathcal{K})|\mathbf{w}_\Omega)$ (Fig.~\ref{fig:volumes}).
\begin{equation}
\label{eq:geo_vol}
    \mathbf{G}=V(\mathbf{C}(\mathcal{K})|\mathbf{w}_\Omega)
\end{equation}
This embedding volume encodes the feature correlations across key views in the video, which have been propagated using downsampling and upsampling layers with skip connections. This structure allows the volume to represent the static appearance elements from the video. In this way, the system can generalise to new scene arrangements and video lengths at inference time.

\paragraph{Motion volume}
This volume is similar in concept to the \emph{geometry volume}, but it extracts dynamic correlations across short-term neighbouring frames instead of modelling static structure. By choosing only frames near our target time, the network is informed of the correlations caused by moving objects (Fig.~\ref{fig:volumes}).

We choose \emph{M} neighbours around our target frame at time $t$ to create this motion encoding volume. In practice, we use $M=4$ for neighbours $\mathcal{N}_t=\{\mathbf{I}_{i}\mid \mathbf{I}_{i} \in \mathcal{V}, i \in \{t\pm1,t\pm2\}\}$. We build the motion volume following the same approach as for the static geometry encoding volume. Using the neighbouring frames $\mathbf{I}_{i} \in \mathcal{N}_t$ and the homography in Equation~\ref{eq:homography} to build the feature sweep volumes $S(\mathbf{I}_{i})$. Then aggregating the feature volumes in the cost volume $\mathbf{C}(\mathcal{N}_t)$. And subsequently, passing the cost volume through a 3D CNN network with the same architecture as for the \emph{geometry volume}, but with different training weights. We then obtain the motion volume,
\begin{equation}
\label{eq:mot_vol}
    \mathbf{M}=V(\mathbf{C}(\mathcal{N}_t)|\mathbf{w}_\Xi)
\end{equation}
This volume thus represents the short-term behaviours of dynamic scene elements.

\subsection{Static neural radiance fields}
We optimise an MLP~\cite{mildenhall2020nerf} $F_{\Theta}$ with parameters $\mathbf{w}_\Theta$ to decode the \emph{geometry volume} embedding into a density and view-dependent radiance (colour). Given a 3D point $x$ and a viewing direction $d$, the network $F_{\Theta}$ regresses the density $\sigma$ and colour $c$ at that point, conditioned on the \emph{geometry volume} $\mathbf{G}$ from Equation~\ref{eq:geo_vol}. To allow the correlations and structures in $\mathbf{G}$ to be mapped back to the original scene albedo, we use the pixel colour of the original key frame inputs $\mathcal{K}$ as additional conditioning information~\cite{chen2021mvsnerf}. We also predict blending weights $b$~\cite{li2021NSFF}, which assign the linear weights to blend the colour and density estimated by the static and dynamic radiance fields (see Sec.~\ref{subsec:volume_rendering}). These weights are unsupervised and give higher importance to the static regions of the scene that this static representation can best model.
\begin{equation}
    F_{\Theta}\colon \left(x,d,\mathbf{G},\mathcal{K}|\mathbf{w}_\Theta\right) \mapsto \left(\sigma_{x}, c_{x,d}, b_{x,d}\right)
\end{equation}
\subsection{Dynamic neural radiance fields}
\label{subsec:dynerf}
To model the dynamics of our scene, we cannot just naively add a time dimension to our radiance field input. Doing so results in very noisy and inconsistent results due to the problem's dimensionality~\cite{li2021NSFF,li2022DyNeRF}. Instead, we estimate scene flow fields~\cite{li2021NSFF} to aggregate information between frames. Our dynamic representation predicts the forward and backward 3D scene flow at a given location $\mathcal{F}_t=\left( \mathbf{f}_{t \rightarrow t+1}, \mathbf{f}_{t \rightarrow t-1}\right)$. This denotes 3D offset vectors that point to the position of that point at times $t+1$ and $t-1$.

Unfortunately, density and scene flow are ambiguous in disocclusion regions caused by 3D motion. There is no simple mechanism to distinguish a region with a low correlation due to being empty and one with a low correlation due to being occluded in all frames. The prevalence of these fully occluded regions grows drastically as the number of input frames is reduced, leading traditional radiance field models to produce reconstructions full of unrealistic holes. To avoid this, we also predict disocclusion weights $\mathcal{W}_t=\left( \mathnormal{w}_{t \rightarrow t+1}, \mathnormal{w}_{t \rightarrow t-1}\right)$. These unsupervised weights can be seen as the confidence of the estimated results and can guide the application strength of the reconstruction losses to the areas we are certain are observable across the video.
The dynamic representation is then given by the following function $F_{\Phi}$
where $x$ is the query point in 3D space, with viewing direction $d$ at time $t \in \{1,...,N\}$. $\mathbf{M}$ is our motion encoding volume from Equation~\ref{eq:mot_vol}, to which we concatenate the original colour values from neighbouring frames $\mathcal{N}_t$.
\begin{equation}
    F_{\Phi}\colon \left(x,d,t,\mathbf{M},\mathcal{N}_t|\mathbf{w}_\Phi\right) \mapsto \left(\sigma_{x,t}, c_{x,d,t}, \mathcal{F}_t, \mathcal{W}_t\right).
\end{equation}
\subsection{Volume rendering}
\label{subsec:volume_rendering}
We use differentiable ray marching to render the colour of image pixels. This is done by projecting (``marching'') a ray $\mathbf{r}_t$ through a pixel in the target image $\mathbf{I}_{t}$ at time $t$. We query the respective neural radiance networks at regular intervals $\gamma \in [1..\inf]$ along this ray to obtain the radiance (colour) $c_{\gamma}=c\left(\gamma\right)$ and density $\sigma_{\gamma}=\sigma\left(\gamma\right)$ at each ray sample.
\begin{equation}
    \left(\sigma_{\gamma}, c_{\gamma}, b_{\gamma}\right) = F_{\Theta}\left(\mathbf{r}_{t,\gamma},\mathbf{d},\mathbf{G},\mathcal{K}|\mathbf{w}_\Theta\right),\;\;
    \left(\sigma_{t,\gamma}, c_{t,\gamma}, \mathcal{F}_{t,\gamma}, \mathcal{W}_{t,\gamma}\right)= F_{\Phi}\left(\mathbf{r}_{t,\gamma},\mathbf{d},t,\mathbf{M},\mathcal{N}_t|\mathbf{w}_\Phi\right)
\end{equation}
where $\mathbf{r}_{t,\gamma}=\mathbf{r}_t(\gamma) = \mathbf{o}_{t} + \gamma\mathbf{d}$ with $\mathbf{o}_{t}$ being the ray origin and $\mathbf{d}$ the direction vector.
Then, we use the predicted blending weights $b_{\gamma}$ to ``blend'' the colour and density samples from $F_{\Theta}$ and $F_{\Phi}$. Obtaining the estimated colour $\hat{C}^{b}(\mathbf{r}_t)$ of the pixel via the volume rendering Equation~\cite{kajiya1984}:
\begin{gather}
    \label{eq:ray_render}
    \hat{C}^{b}_{sta}(\mathbf{r}_{t}) = \sum_{\gamma} \tau^{b}_{\gamma} \left(1-exp\left(-\sigma_\gamma\right)\right)c_\gamma,
    \;\;\;\;\;\;\;\;\;\;\;\;\;\;\;\;\hat{C}^{b}_{dy}(\mathbf{r}_{t}) = \sum_{\gamma} \tau^{b}_\gamma \left(1-exp\left(-\sigma_{t,\gamma}\right)\right)c_{t,\gamma} 
    \\
    \hat{C}^{b}(\mathbf{r}_{t}) = \left(1-b_\gamma \right) \hat{C}^{b}_{sta}(\mathbf{r}_{t}) + b_\gamma \hat{C}^{b}_{dy}(\mathbf{r}_{t})
\end{gather}
where $\tau^{b}_\gamma$ is the blended transmittance at sample $\gamma$, which represents the probability that the ray travels up to $\gamma$ without hitting another particle.
\begin{equation}
    \tau^{b}_\gamma = exp\left(-\sum_{j=1}^{\gamma-1} \left( \sigma_j \left(1-b_j \right) + \sigma_{t,j} b_j \right) \right)
\end{equation}
\subsection{Losses and regularisations}
\paragraph{Reconstruction loss}
To enforce low-frequency correctness in the output, we compute the $L_2$ loss 
between the blended colour estimate and the true colour,
\begin{equation}
    \mathcal{L}_{rec} = ||\hat{C}^{b}(\mathbf{r}_{t})-C(\mathbf{r}_{t})||^2_2
\end{equation}

\paragraph{Temporal photometric consistency}
To encourage the reconstruction of the scene at time $t$ to be consistent with the scene at neighbouring times $k \in \mathcal{N}\left(t\right)$. We apply a photometric loss~\cite{li2021NSFF} to our dynamic network, which considers the motion due to 3D scene flow. This is done by warping the scene from neighbouring frame $k$ to time $t$ and comparing it to the ground truth scene at time $t$. Because this loss suffers from ambiguity in disocclusion areas, it is scaled by the estimated confidence weights mentioned in Sec.~\ref{subsec:dynerf}.
\begin{equation}
    \mathcal{L}_{pho} = \sum_{k \in \mathcal{N}\left(t\right) 
    %\cup \left\{t\right\}
    } \hat{W}_{k \rightarrow t}(\mathbf{r}_{t})||\hat{C}_{dy}(\mathbf{r}_{t} + \mathbf{f}_{t \rightarrow k})-C(\mathbf{r}_{t})||^2_2
\end{equation}
where $\mathbf{f}_{t \rightarrow k} \in \mathcal{F}_{t}$ is the scene flow field from time $t$ to $k$. We apply volume rendering to get the colour $\hat{C}_{dy}(\mathbf{r}_{t} + \mathbf{f}_{t \rightarrow k})$ from the dynamic network $F_{\Phi}$ at time $k$. We also apply volume rendering to get the disocclusion weights $\hat{W}_{k \rightarrow t}(\mathbf{r}_{t})$, which are the accumulated confidence estimates $w_{t \rightarrow k, \gamma}$ along each ray.
\begin{align}
    \hat{C}_{dy}(\mathbf{r}_{t} + \mathbf{f}_{t \rightarrow k}) &= \sum_{\gamma} \tau_{k,\gamma} \left(1-exp\left(-\sigma_{k,\gamma}\right)\right)c_{k,\gamma},
    \;\;\;\;\;\;\;\;\;\;\tau_{k,\gamma} = exp\left(-\sum_{j=1}^{\gamma-1} \left( \sigma_{k,j} \right) \right)
    \\
    \hat{W}_{k \rightarrow t}(\mathbf{r}_{t}) &= \sum_{\gamma} \tau_{k,\gamma} \left(1-exp\left(-\sigma_{k,\gamma}\right)\right)\mathnormal{w}_{t \rightarrow k, \gamma}
\end{align}

\paragraph{Disocclusion weight regularisation}
The system can fall into a degenerate local minimum, where the disocclusion weights $w_{t \rightarrow k}$ are zero at every 3D point $x_t$, and only the static NeRF is active.  We avoid this with an $L_1$ regularisation that pushes them closer to one.
\begin{equation}
    \mathcal{L}_{\mathnormal{w}} = \sum_{x_t} \sum_{k \in \mathcal{N}\left(t\right)}||\mathnormal{w}_{t \rightarrow k}(x_t)-1||
\end{equation}

\paragraph{Blending weights entropy loss}
This loss encourages blending weight to be either 0 or 1, which can help to reduce the ghosting caused by learned semi-transparent blending weights.
\begin{equation}
    \mathcal{L}_{\mathnormal{b}} = ||-b \cdot log\left(b\right)||
\end{equation}

\paragraph{Cycle loss}
We enforce a cyclic regularisation to ensure that the predicted forward flow field at time $t$ ($\mathbf{f}_{t \rightarrow k}$) is consistent with the backward flow field at time $k$ ($\mathbf{f}_{k \rightarrow t}$), where $k$ is a neighbouring time $k \in \left\{t \pm 1\right\}$. Fundamentally, if $x_{t \rightarrow k} = x_{t} + \mathbf{f}_{t \rightarrow k}$ then we should see that $x_{t \rightarrow k} + \mathbf{f}_{k \rightarrow t} = x_{t}$. That is, $\mathbf{f}_{t \rightarrow k} = -\mathbf{f}_{k \rightarrow t}$. Flow fields are ambiguous at disocclusion areas, so we regulate this loss with the predicted disocclusion weights~\cite{li2021NSFF}.
\begin{equation}
    \mathcal{L}_{cyc} = \sum_{x_t}\sum_{k \in \{t \pm 1\}}w_{t \rightarrow k}||\mathbf{f}_{t \rightarrow k}(x_t) +\mathbf{f}_{k \rightarrow t}(x_{t \rightarrow k})||
\end{equation}

\paragraph{Scene flow regularisation}
We assume that motion is small in most 3D space~\cite{valmadre2012}, so we minimise the absolute value of the flow fields.
\begin{equation}
    \mathcal{L}_{min} = \sum_{x_t}\sum_{k \in \{t \pm 1\}}||\mathbf{f}_{t \rightarrow k}(x_t)||
\end{equation}

\paragraph{Scene flow spatial smoothness}
We assume that the scene deforms in a piece-wise smooth way~\cite{newcombe2015dynamicfusion}, meaning the scene flow field is spatially smooth.
\begin{equation}
    \mathcal{L}_{sp} = \sum_{x_t}\sum_{y_t \in \mathcal{N}\left(x_t\right)}\sum_{k \in \{t \pm 1\}}\mathrm{w}^{dist}\left(x_t,y_t\right)||\mathbf{f}_{t \rightarrow k}(x_t) -\mathbf{f}_{t \rightarrow k}(y_t)||
\end{equation}
where $\mathcal{N}\left(x_t\right)$ are the neighbouring points of $x_t$, and $\mathrm{w}^{dist}\left(x_t,y_t\right) = exp(-2||x_t-y_t||)$ are weightings based on Euclidean distance.
\paragraph{Scene flow temporal smoothness}
Finally, we add a scene flow temporal smoothness loss. This loss encourages 3D point trajectories to have minimal kinetic energy~\cite{vo2016} i.e. constant velocity and piece-wise linear motion.
\begin{equation}
    \mathcal{L}_{temp} = \sum_{x_t}||\mathbf{f}_{t \rightarrow t+1}(x_t) +\mathbf{f}_{t \rightarrow t-1}(x_t)||^2_2
\end{equation}

\subsection{Weakly-supervised pre-training}
Since the problem of reconstructing complex dynamic scenes is extremely ill-posed, the losses can converge to local minima when randomly initialised~\cite{li2021NSFF}. We first complete a data-mining stage to initialise the problem optimally. We use monocular optical flow and depth estimation networks to generate pseudo ground truth data to guide our network. However, as both models are not completely accurate, we use them for initialisation only and decay their contribution to zero during training.

\paragraph{Geometric consistency}
First, we compute a reprojection error between the scene flow field estimated by ZeST-NeRF and that derived from pre-trained optical flow models~\cite{teed2020RAFT}. 
Given a ray $\mathbf{r}_{t}$ at time $t$ and estimated 3D scene-flow $\mathbf{f}_{t \rightarrow k}$, we can calculate the expected location of the 3D point $\hat{X}_{t}(\mathbf{r}_{t})$ and the expected 2D optical flow $\hat{F}_{t \rightarrow k}(\mathbf{r}_{t})$ of that point, by using volume rendering,
\begin{equation}
    \hat{X}_{t}(\mathbf{r}_{t}) = \sum_{\gamma} \tau_{t,\gamma} \left(1-exp\left(-\sigma_{t,\gamma}\right)\right)x_{t,\gamma}, \;\;\;\;\;\;
    \hat{F}_{t \rightarrow k}(\mathbf{r}_{t}) = \sum_{\gamma} \tau_{t,\gamma} \left(1-exp\left(-\sigma_{t,\gamma}\right)\right)\mathbf{f}_{t \rightarrow k, \gamma}
\end{equation}
Then we project that displaced 3D point $\hat{X}_{t}(\mathbf{r}_{t}) + \hat{F}_{t \rightarrow k}(\mathbf{r}_{t})$ into the camera view of the neighbouring frame ($k$) as $\hat{p}_{t \rightarrow k} = \pi\left(\mathbf{K}\left(\mathbf{R}_{k}\left(\hat{X}_{t}(\mathbf{r}_{t}) + \hat{F}_{t \rightarrow k}(\mathbf{r}_{t})\right) + \mathbf{t}_{k}\right)\right)$.
We compare our estimation to the displaced pixel $p_{t \rightarrow k}=p_{t}+\mathbf{u}_{t \rightarrow k}$, where $\mathbf{u}_{t \rightarrow k}$ is the 2D optical flow derived using a pre-trained optical flow model~\cite{teed2020RAFT}. We can then minimise the reprojection error with the following loss~\cite{innmann2020NRMVS},
\begin{equation}
    \mathcal{L}_{geo} = \sum_{k \in \{t \pm 1\}}||\hat{p}_{t \rightarrow k} - p_{t \rightarrow k}||
\end{equation}

\paragraph{Single-view depth prior}
In traditional \textit{neural radiance field} approaches, depth maps are extremely noisy, if not nonsensical. We use a pre-trained monocular depth estimation network~\cite{ranftl2022midas} to  encourage the expected termination depth along each ray $\hat{D}(\mathbf{r}_{t})$ to be close to the derived depth $D(\mathbf{r}_{t})$. We apply an $L_1$ loss using a robust scale-shift invariant metric~\cite{li2021NSFF},
\begin{equation}
\mathcal{L}_{depth} = ||\hat{D}(\mathbf{r}_{t})-D(\mathbf{r}_{t})||.
\end{equation}
%-------------------------------------------------------------------------
\section{Evaluation}
\label{sec:experiments}

\paragraph{Implementation details}
We use the same architecture to build both the \emph{geometry} and \emph{motion} volumes, but they are optimised separately and thus do not share weights. We extract 32 feature channels from the input images using a 2D CNN consisting of downsampling convolutional layers with Batch-Normalisation (BN) and ReLU activation function. Warping these features we create a Plane Sweep Volume per input image, selecting $D=128$ depth planes. These are aggregated through a variance-based cost volume and processed using a 3D UNet-like network, consisting of downsampling and upsampling Convolutional layers with BN and ReLU, and skip connections.

For the NeRF MLPs, we follow a similar setup to the original case~\cite{mildenhall2020nerf}. We sample 128 points along each ray, with a ray batch of 1024. We also have two separate networks for the static and dynamic parts, which do not share weights. We append the normalised time indices in NSFF~\cite{li2021NSFF} to our dynamic network inputs. The MLP networks return the estimated colour $c$ and density $\sigma$, as well as blending weights $b$ in the case of the Static MLP, and 3D scene flow $\mathrm{f}$ and occlusion weights $w$ in the case of the Dynamic MLP. We use an Adam optimiser~\cite{kingma2014adam} with a learning rate of $5e-4$. We use positional encoding (PE)~\cite{mildenhall2020nerf} for the 3D location and viewing direction before feeding them into the networks. For more detailed information about the architecture, refer to the supplementary material.

\begin{table}[htpb]
    \captionsetup{skip=3pt,width=.9\textwidth}
	\caption{\textbf{Quantitative evaluation.} Results on cross-validation training and further fine-tuning.  \textbf{Bold} is best result, \textit{italic} is second best.}
	\centering
	\begin{tabular}{ lCCC@{\hskip 0.3in}CCC } 
 
    	\toprule
         \multirow{2}{6em}{Model} & \multicolumn{3}{c}{Scene-agnostic} & \multicolumn{3}{c}{Scene-specific}\\%[2pt]

         \cmidrule(l{.5em}r{1.3em}){2-4}
         \cmidrule{5-7}
    	\rule{0pt}{3ex} & PSNR$\uparrow$ & SSIM$\uparrow$ & LPIPS$\downarrow$ & PSNR$\uparrow$ & SSIM$\uparrow$ & LPIPS$\downarrow$\\
        \midrule
            NSFF~\cite{li2021NSFF} & 13.15 & 0.3178 & 0.7334 & \textbf{28.19} & \textbf{0.9280} & \textbf{0.0450}  \\ 
            SVS~\cite{menendez2022svs}  & 17.91 & 0.4958 & 0.3698 & 21.28 & 0.7103	& 0.2001  \\
            MVSNeRF~\cite{chen2021mvsnerf} & \textit{19.37} & \textit{0.6198} & \textit{0.2885} & \textit{26.94}	& \textit{0.8678} & 0.1208\\ 
            \midrule
            ZeST-NeRF (Ours)  & \textbf{21.59} & \textbf{0.6239}  & \textbf{0.2048} & \textit{26.94}	& 0.8575 & \textit{0.0995}  \\ 
        \bottomrule
	\end{tabular}
	\label{table:quantitative_evaluation}
\end{table}

\paragraph{Baseline models and datasets}
We compare our results to a space-time view synthesis approach NSFF~\cite{li2021NSFF} and two scene-agnostic static view synthesis approaches SVS~\cite{menendez2022svs} and MVSNeRF~\cite{chen2021mvsnerf} applied naively across frames. We attempted to reproduce the results in Li~\etal~\cite{li2022DyNeRF} but found it impossible with the information given in the paper due to the lack of published code.
We train on the Dynamic Scenes dataset~\cite{yoon2020DFNet}. This dataset consists of 8 short-time video clips recorded with 12 synchronised cameras, and sampled from each camera at different times to simulate a moving monocular camera. The scenes are collected in the wild and feature complex motions such as jumping, running, or dancing. As this dataset is quite small, we perform a leave-one-out cross-validation study to prove our performance. We train one full model per subset of scenes, created by holding out one single scene for validation. This means that we can test on every scene of the Dynamic Scenes dataset with a model that has been trained on completely unrelated scenes.
Please refer to the supplementary material for additional details regarding the experimental setup.

\newcommand\imwidth{0.19\textwidth}
\begin{figure}[!htb]
    \centering
    \setlength{\tabcolsep}{1pt}
    \begin{tabular}{ccccc}
     NSFF~\cite{li2021NSFF} & SVS~\cite{menendez2022svs} & MVSNeRF~\cite{chen2021mvsnerf} & {\small ZeST-NeRF (Ours)} & Ground Truth\\
     
     \includegraphics[width=2.5cm,height=1.4cm]{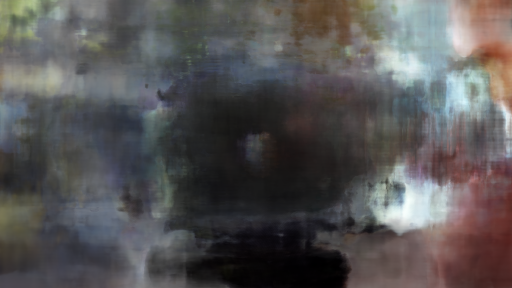} & \includegraphics[width=2.5cm,height=1.4cm]{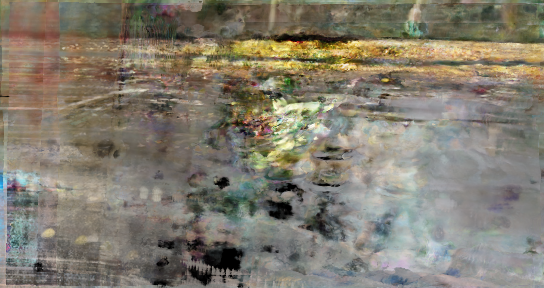} & \includegraphics[width=2.5cm,height=1.4cm]{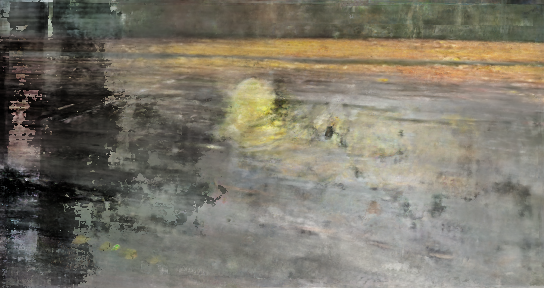} & \includegraphics[width=2.5cm,height=1.4cm]{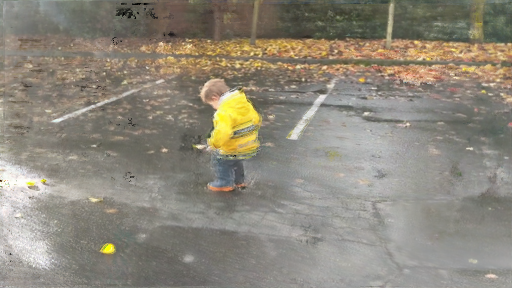} & \includegraphics[width=2.5cm,height=1.4cm]{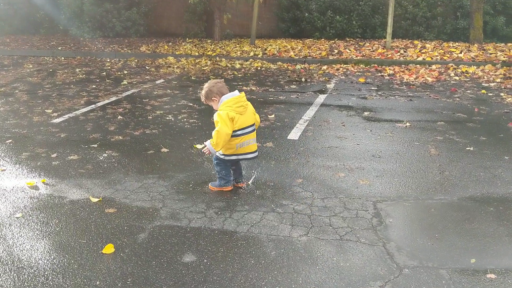}\\
     
     \includegraphics[width=2.5cm,height=1.4cm]{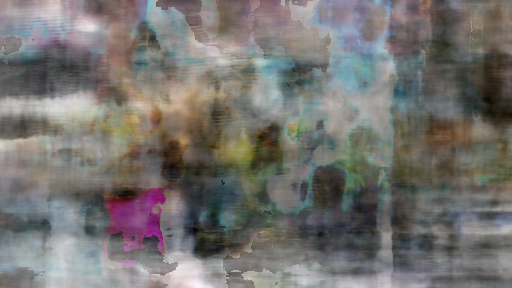} & \includegraphics[width=2.5cm,height=1.4cm]{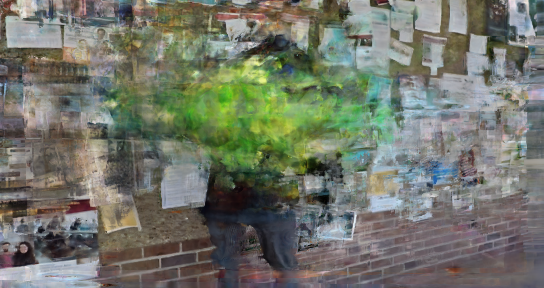} & \includegraphics[width=2.5cm,height=1.4cm]{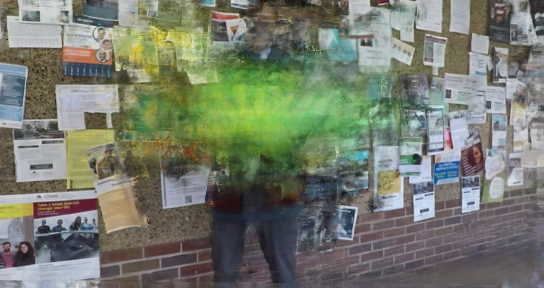} & \includegraphics[width=2.5cm,height=1.4cm]{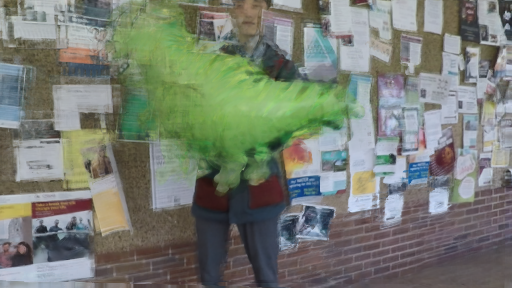} &  \includegraphics[width=2.5cm,height=1.4cm]{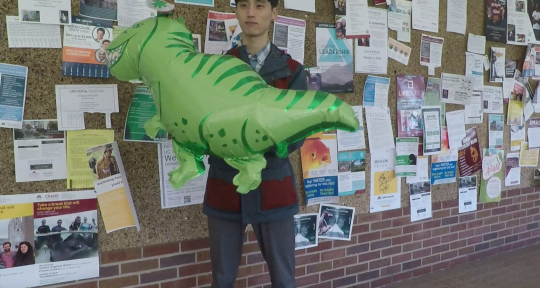}\\
     
     \includegraphics[width=2.5cm,height=1.4cm]{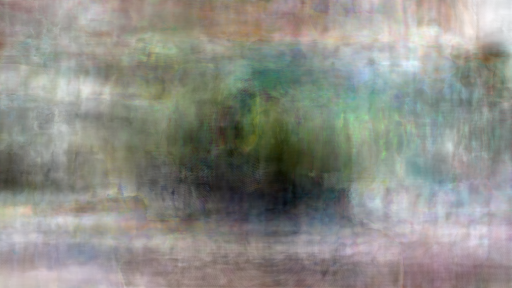} & \includegraphics[width=2.5cm,height=1.4cm]{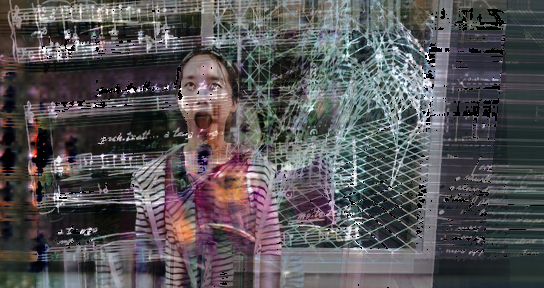} & \includegraphics[width=2.5cm,height=1.4cm]{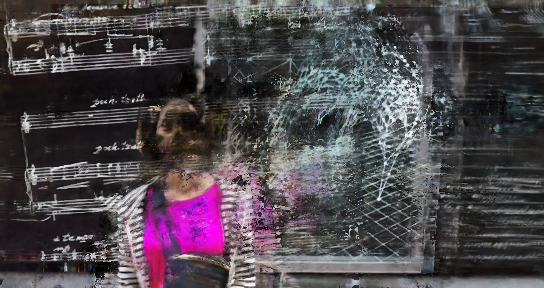} & \includegraphics[width=2.5cm,height=1.4cm]{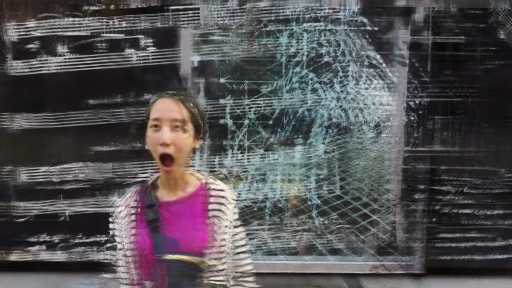} &
     \includegraphics[width=2.5cm,height=1.4cm]{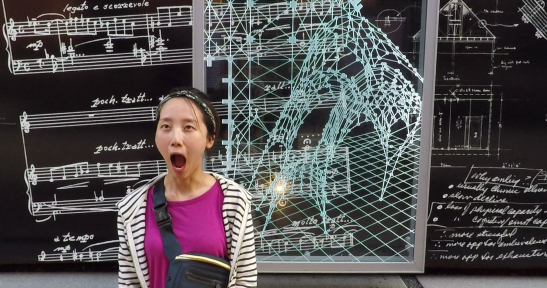}\\
     
     \includegraphics[width=2.5cm,height=1.4cm]{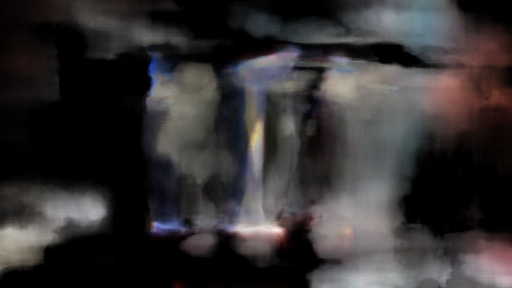} & \includegraphics[width=2.5cm,height=1.4cm]{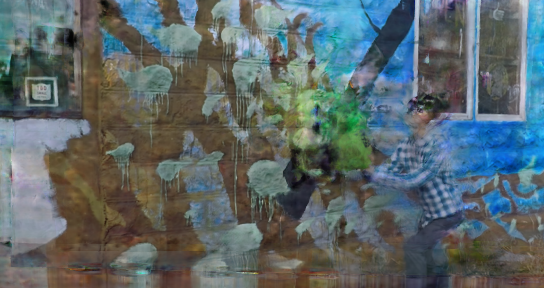} & \includegraphics[width=2.5cm,height=1.4cm]{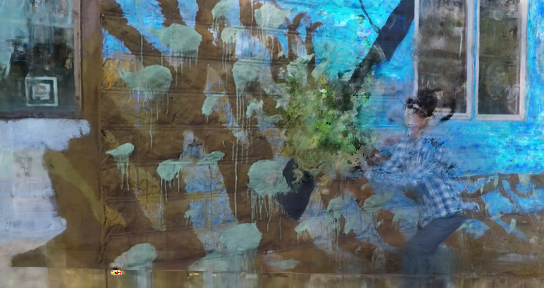} & \includegraphics[width=2.5cm,height=1.4cm]{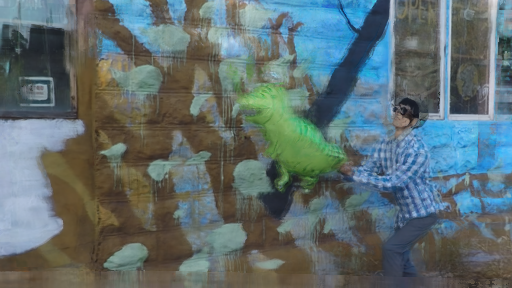} & \includegraphics[width=2.5cm,height=1.4cm]{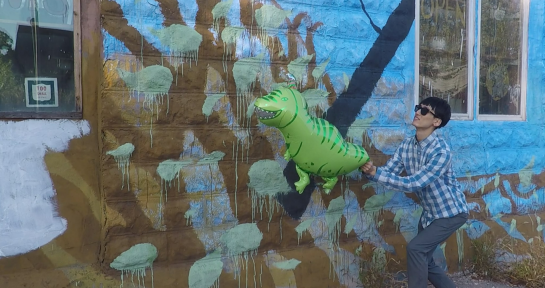}\\
     
     \includegraphics[width=2.5cm,height=1.4cm]{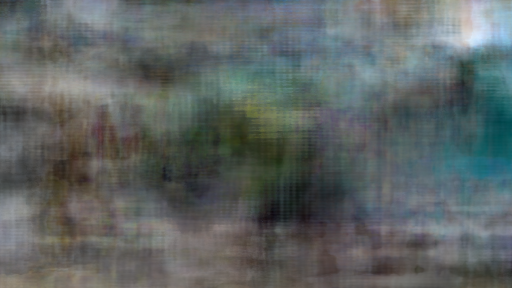} & \includegraphics[width=2.5cm,height=1.4cm]{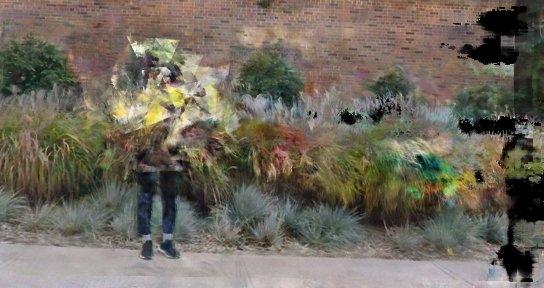} & \includegraphics[width=2.5cm,height=1.4cm]{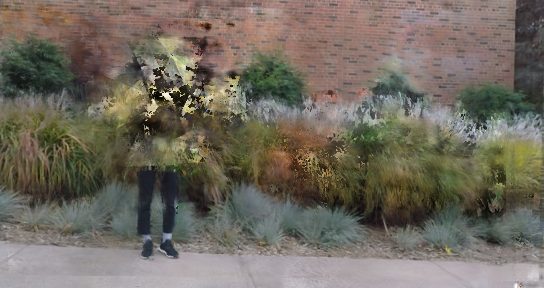} & \includegraphics[width=2.5cm,height=1.4cm]{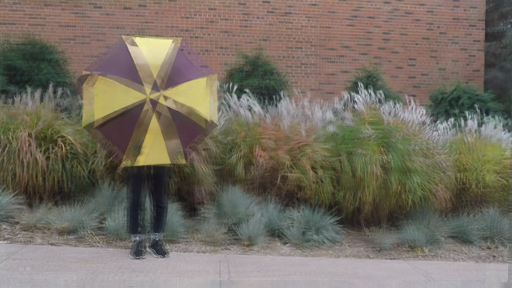} &     \includegraphics[width=2.5cm,height=1.4cm]{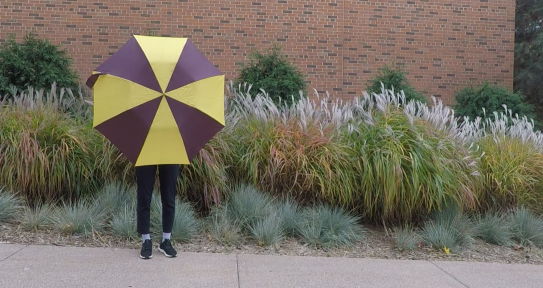} \\
    \end{tabular}
    \vspace{0.4cm}
    \caption{\textbf{Qualitative results:} Example results (see supplementary material for more).}
    \label{fig:qualitative}
\end{figure}

\paragraph{Results}
When comparing ZeST-NeRF to other state-of-the-art models in Table~\ref{table:quantitative_evaluation}, we can see that our approach outperforms all models when applied in a scene-agnostic manner. ZeST-NeRF improves the results of other scene-agnostic methods by at least 15\% across all metrics. In addition, a ZeST-NeRF model pre-trained on different sequences can be rapidly fine-tuned on a specific scene to produce results competitive with exhaustively optimised scene-specific approaches quickly. Qualitative results are shown in Figure~\ref{fig:qualitative} (see supplementary material for additional qualitative examples). We can observe that our model produces significantly more accurate and pleasant results than the other methods. Approaches like MVSNeRF~\cite{chen2021mvsnerf} or SVS~\cite{menendez2022svs} are able to reconstruct the correct background, but struggle to recover the dynamic objects in the scene. As the numbers from the ablation study suggest (Table~\ref{table:ablation_study}), this is probably due to the necessity of having both a Geometry and a Dynamic network to recover static and dynamic features. On the other hand, the NSFF~\cite{li2021NSFF} model is not really equipped to solve this scene-agnostic problem.

Despite the inherent complexity and ill-posed nature of the scene-agnostic ``zero-shot'' novel view synthesis problem, it is important to acknowledge that our approach is not without its limitations.
As the network concentrates on recovering dynamic areas, the background reconstruction suffers from a slight quality reduction and some blurriness may persist in certain scenes.
Furthermore, our dynamic reconstruction occasionally suffers from duplication artefacts.
Despite this, our approach is a significant step-change in performance compared to previous research.

\begin{table}[htpb]
    \captionsetup{skip=3pt,width=.9\linewidth}
	\caption{\textbf{Ablation study.} Effect of each addition to the model. \textbf{Bold} is best result.}
	\centering
	\begin{tabular}{ ccccc }
    	\toprule
    	Geometry & Dynamic  & PSNR$\uparrow$ & SSIM$\uparrow$ & LPIPS$\downarrow$\\
    	\midrule
    	$\times$ & $\times$ & 12.67 & 0.4142 & 0.8277 \\ 
    	$\checkmark$ & $\times$  & 17.87 & 0.4533 & 0.4781 \\ 
            $\times$ & $\checkmark$  & {20.19} & {0.5814} & {0.2447}\\ 
    	$\checkmark$ & $\checkmark$ & \textbf{21.59} & \textbf{0.6239}  & \textbf{0.2048} \\ 
    	\bottomrule
	\end{tabular}
	\label{table:ablation_study}
\vspace{-0.4cm}
\end{table}

\section{Conclusions}
In this paper we have proposed ZeST-NeRF. This dynamic-scene representation technique combines multi-view synthesis and scene flow field estimation approaches to rendering novel views in previously unseen scenes. This is useful for applications in media production, where we have many videos and want to generate new camera views without expensive scene-specific training.

Attempting to apply previous state-of-the-art techniques from related fields naively, leads to blurry results with overwhelming numbers of artefacts or, in some cases, a complete inability to train.
The problem of scene-agnostic ``zero-shot'' novel view synthesis is highly ill-posed, and our model exhibits certain limitations.
Artefacts like duplications persist in certain scenes, particularly in regions characterised by temporal motion.
In addition, some blurriness may occur in background areas.
Furthermore, given the limited training dataset, our generalisation power may be constrained by the type of scene dynamics and length of motion. It would be interesting to study the behaviour of the model in larger datasets with a more varied set of motions and scene complexities.
Nonetheless, our approach constitutes a significant advancement in performance compared to prior research efforts.

In the future, it would be interesting to explore techniques that can better reconstruct dynamic objects, potentially aided by stereo views from multiple cameras. Other representation techniques, like instant Neural Graphic Primitives~\cite{muller2022NGP}, may help improve efficiency. In addition, generative models like diffusion models could be explored to tackle uncertainty in highly occluded areas or areas with more significant motion.

\paragraph{Acknowledgements.}This work was partially supported by the British Broadcasting Corporation (BBC) and the Engineering and Physical Sciences Research Council's (EPSRC) industrial CASE project ``Generating virtual camera views with generative networks'' (voucher number 19000033).

\bibliography{vua_library}

\newpage
\appendix
% \begin{appendices}
\section{Implementation details}
We use COLMAP~\cite{schonberger2016COLMAP} to generate camera intrinsics and extrinsics at each frame while masking features from regions associated with dynamic objects~\cite{li2021NSFF} using off-the-shelf instance segmentation~\cite{he2017maskrcnn}.
We extract deep image features from the selected frames using a 2D CNN network with 32 channels (first section of Table~\ref{tbl:architecture_vol}). These features are used to construct the plane sweep volume~\cite{flynn2016deepstereo} using 128 depth planes. These sweep volumes are then aggregated into a variance-based cost volume. This is then processed into the \textit{geometry} and \textit{motion} volumes as defined by the 3D CNN architecture on the second section of Table~\ref{tbl:architecture_vol}. These volumes have the same architecture, only differing in the number of input channels ($K=8$ key-frames and $N=4$ neighbours, respectively). The \textit{geometry} and \textit{motion} volumes do not share their weights.

\begin{table}[htp]
\captionsetup{skip=3pt,width=.9\textwidth}
\caption{\textbf{Encoding volumes architecture}: g/m denote the geometry and motion 3D features respectively.
\textbf{k} is the kernel size, \textbf{s} is the stride, \textbf{d} is the kernel dilation, and \textbf{chns} shows the number of input and output channels for each layer.
We denote CBR2D/CBR3D/CTB3D to be ConvBnReLU2D, ConvBnReLU3D, and ConvTransposeBn3D layer structure respectively.}
% \vspace{0.5cm}
    \center
    \begin{tabular}{ccccccc}
    \toprule\hline
    & \textbf{Layer}  & \textbf{k} & \textbf{s} & \textbf{d}  & \textbf{chns} & input  \\ \hline
    \multirow{8}{*}{2D CNN} & CBR2D$_0$ & 3 & 1 & 1 & $3/8$    & $I$\\
    & CBR2D$_1$ & 3 & 1 & 1 & $8/8$    & CBR2D$_0$\\
    & CBR2D$_2$ & 5 & 2 & 2 & $8/16$   & CBR2D$_1$\\
    & CBR2D$_3$ & 3 & 1 & 1 & $16/16$  & CBR2D$_2$\\
    & CBR2D$_4$ & 3 & 1 & 1 & $16/16$  & CBR2D$_3$\\
    & CBR2D$_5$ & 5 & 2 & 2 & $16/32$  & CBR2D$_4$\\
    & CBR2D$_6$ & 3 & 1 & 1 & $32/32$  & CBR2D$_5$\\
    & $E$ = CBR2D$_7$      & 3 & 1 & 1 & $32/32$  & CBR2D$_6$\\ \hline

    \multirow{11}{*}{3D CNN} & CBR3D$_0$ & 3 & 1 & 1 & $32+(K/N)*3/8$  & $E, I$\\
    & CBR3D$_1$ & 3 & 2 & 1 & $8/16$  & CBR3D$_0$\\
    & CBR3D$_2$ & 3 & 1 & 1 & $16/16$ & CBR3D$_1$\\
    & CBR3D$_3$ & 3 & 2 & 1 & $16/32$  & CBR3D$_2$\\
    & CBR3D$_4$ & 3 & 1 & 1 & $32/32$ & CBR3D$_3$\\
    & CBR3D$_5$ & 3 & 2 & 1 & $32/64$  & CBR3D$_4$\\
    & CBR3D$_6$ & 3 & 1 & 1 & $64/64$ & CBR3D$_5$\\
    & CTB3D$_0$ & 3 & 2 & 1 & $64/32$ & CBR3D$_6$\\
    & CTB3D$_1$ & 3 & 2 & 1 & $64/32$ & CTB3D$_0$ + CBR3D$_4$\\
    & CTB3D$_2$ & 3 & 2 & 1 & $64/32$ & CTB3D$_1$ + CBR3D$_2$\\
    & $g/m$ = CTB3D$_3$ & 3 & 2 & 1 & $64/32$ & CTB3D$_2$ + CBR3D$_0$\\

    \bottomrule
    \end{tabular}
% \vspace{0.2cm}
\label{tbl:architecture_vol}
\end{table}

For the NeRF MLPs, we follow a similar setup to the original case~\cite{mildenhall2020nerf}. We sample 128 points along each ray, with a ray batch of 1024. We also have two separate networks for the static and dynamic parts, which do not share weights. We append the normalised time indices in NSFF~\cite{li2021NSFF} to our dynamic network inputs. The MLP networks return the estimated colour $c$ and density $\sigma$, as well as blending weights $b$ in the case of the Static MLP, and 3D scene flow $\mathrm{f}$ and occlusion weights $w$ in the case of the Dynamic MLP. We use an Adam optimiser~\cite{kingma2014adam} with a learning rate of $5e-4$. We use positional encoding (PE)~\cite{mildenhall2020nerf} for the 3D location and viewing direction before feeding them into the networks. For more detailed information about the architecture, refer to the Table~\ref{tbl:architecture_mlp}.

\begin{table}[htbp]
\captionsetup{skip=3pt,width=.9\textwidth}
\caption{\textbf{MLPs architecture}: g/m denote the geometry and motion 3D features respectively.
k and n are the original colours of the K key-frames and N neighbouring frames, that are concatenated to the inputs.
\textbf{chns} shows the number of input and output channels for each layer.
We denote LR to be LinearReLU layer structure.
PE refers to the positional encoding as used in \cite{mildenhall2020nerf}.}
% \vspace{0.5cm}
    \center
    \begin{tabular}{cccc}
    \toprule\hline
    & \textbf{Layer}  & \textbf{chns} & input  \\ \hline

    \multirow{9}{*}{Static MLP} & PE$_0$  & 3/63 & $x$ \\
    & LR$_0$ & 8+K*3/256 & $g,k$ \\
    & LR$_1$ & 63/256 & PE\\
    & LR$_{i+1}$ & 256/256 & LR$_i$+LR$_0$ \\
    & $\sigma$ & 256/1 & LR$_6$\\
    & $b$ & 256/1 & LR$_6$\\
    & PE$_1$ & 3/27 & $d$\\
    & LR$_7$ & 27+256/256 & PE$_1$,LR$_6$\\
    & $c$ & 256/3 & LR$_7$\\
    \hline

    \multirow{10}{*}{Temporal MLP} & PE$_0$ & 4/63 & $x,t$ \\
    & LR$_0$ & 8+N*3/256 & $m,n$ \\
    & LR$_1$ & 63/256 & PE\\
    & LR$_{i+1}$ & 256/256 & LR$_i$+LR$_0$ \\
    & $\sigma$ & 256/1 & LR$_6$\\
    & $\mathrm{f}$ & 256/6 & LR$_6$\\
    & $w$ & 256/2 & LR$_6$\\
    & PE$_1$ & 3/27 & $d$\\
    & LR$_7$ & 27+256/256 & PE$_1$,LR$_6$\\
    & $c$ & 256/3 & LR$_7$\\

    \bottomrule
    \end{tabular}
% \vspace{0.2cm}
\label{tbl:architecture_mlp}
\end{table}

\FloatBarrier
% \pagebreak

\section{Evaluation of accuracy}
In order to assess the performance of our model, we employ a range of widely recognized metrics that evaluate various aspects of an image. To measure image quality we make use of the Peak Signal-To-Noise Ratio (PSNR)~\cite{huynh-thu2008psnr} and the Structural SIMilarity (SSIM)~\cite{wang2004ssim} index. PSNR serves as an indicator of the overall consistency of pixels, while SSIM gauges the coherency of local structures.
We define PSNR as
\begin{align}
    \mathit{PSNR} = 10 \cdot \log_{10} \left( \frac{\mathit{MAX}_C^2}{\mathit{MSE}\left(\hat{C}^{b}(\mathbf{r}),C(\mathbf{r})\right) } \right)\\ 
    \mathit{MSE}\left(\hat{C}^{b}(\mathbf{r}),C(\mathbf{r})\right) = \frac{1}{N}\sum_{\mathbf{r}} [\hat{C}^{b}(\mathbf{r})-C(\mathbf{r})]^2
\end{align}
where $\mathit{MAX}_C$ is the maximum possible input value, and $\mathit{MSE}\left(\hat{C}^{b}(\mathbf{r}),C(\mathbf{r})\right)$ represents the per-pixel Maximum Squared Error between the predicted colour $\hat{C}^{b}(\mathbf{r})$ at ray  $\mathbf{r}$, and the original colour $C(\mathbf{r})$, in a batch of N rays.

On the other hand, SSIM is given by
\begin{align}
    \mathit{SSIM}(\hat{C}^{b},C)=\dfrac{(2\mu_{\hat{C}^{b}}\mu_C + k_1)(2\sigma_{\hat{C}^{b}}\sigma_C + k_2)}{(\mu^2_{\hat{C}^{b}} +\mu^2_C + k_1)(\sigma^2_{\hat{C}^{b}}+\sigma^2_C + k_2)}
\end{align}
where $k_1=0.01^2$ and $k_2=0.03^2$ are variables to stabilise the operation. We use a window size of 5 for the Gaussian kernel to smooth the images.

It is worth noting that these metrics assume independence among pixels, which can result in favourable scores for visually inaccurate outcomes. Consequently, we also incorporate the application of a Learned Perceptual Image Patch Similarity (LPIPS)~\cite{zhang2018lpips} metric, which endeavours to capture human perception by leveraging deep features. We use the default settings for the implementation based on AlexNet~\cite{krizhevsky2012alexnet}.

\begin{figure}[htpb]
    \centering
    \setlength{\tabcolsep}{1pt}
    \begin{tabular}{ccccc}
     NSFF~\cite{li2021NSFF} & SVS~\cite{menendez2022svs} & MVSNeRF~\cite{chen2021mvsnerf} & {\small ZeST-NeRF (Ours)} & Ground Truth\\

     \includegraphics[width=2.5cm,height=1.4cm]{images/nsff_media_images_val_rgb_map_blend_69232_1ffc8fe4eb08ce1761d0.png} & \includegraphics[width=2.5cm,height=1.4cm]{images/svs/kid/rgb_map_54.png} & \includegraphics[width=2.5cm,height=1.4cm]{images/mvsnerf/kid/rgb_map_54.png} & \includegraphics[width=2.5cm,height=1.4cm]{images/zest/kid/rgb_map_blend_54.png} & \includegraphics[width=2.5cm,height=1.4cm]{images/gt/kid/00054.png}\\

     \includegraphics[width=2.5cm,height=1.4cm]{images/nsff_baloon_media_images_val_rgb_map_blend_31555_253c5a5cdba63142ccc0.png} & \includegraphics[width=2.5cm,height=1.4cm]{images/svs/balloon1/rgb_map_17.png} & \includegraphics[width=2.5cm,height=1.4cm]{images/mvsnerf/balloon1/rgb_map_17.png} & \includegraphics[width=2.5cm,height=1.4cm]{images/zest/balloon1/rgb_map_blend_17.png} &  \includegraphics[width=2.5cm,height=1.4cm]{images/gt/balloon1/00017.png}\\

     \includegraphics[width=2.5cm,height=1.4cm]{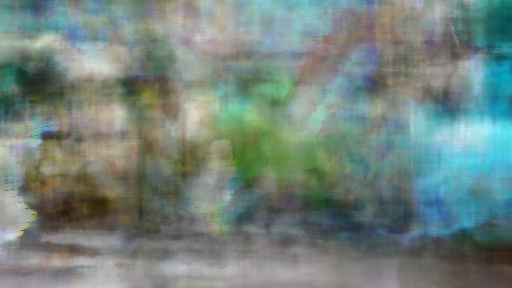} & \includegraphics[width=2.5cm,height=1.4cm]{images/svs/dynamicface/rgb_map_13.png} & \includegraphics[width=2.5cm,height=1.4cm]{images/mvsnerf/dynamicface/rgb_map_13.png} & \includegraphics[width=2.5cm,height=1.4cm]{images/zest/dynamicface/rgb_map_blend_13.png} &
     \includegraphics[width=2.5cm,height=1.4cm]{images/gt/dynamicface/00013.png}\\

     \includegraphics[width=2.5cm,height=1.4cm]{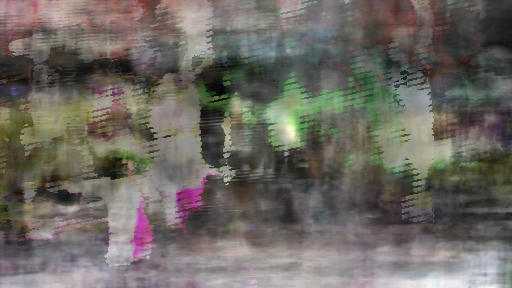} & \includegraphics[width=2.5cm,height=1.4cm]{images/svs/balloon2/rgb_map_10.png} & \includegraphics[width=2.5cm,height=1.4cm]{images/mvsnerf/balloon2/rgb_map_10.png} & \includegraphics[width=2.5cm,height=1.4cm]{images/zest/balloon2/rgb_map_blend_10.png} & \includegraphics[width=2.5cm,height=1.4cm]{images/gt/balloon2/00010.png}\\

     \includegraphics[width=2.5cm,height=1.4cm]{images/nsff_umbrella_media_images_val_rgb_map_blend_19831_2120d13d7ffb53579a50.png} & \includegraphics[width=2.5cm,height=1.4cm]{images/svs/umbrella/rgb_map_19.png} & \includegraphics[width=2.5cm,height=1.4cm]{images/mvsnerf/umbrella/rgb_map_19.png} & \includegraphics[width=2.5cm,height=1.4cm]{images/zest/umbrella/rgb_map_blend_19.png} &     \includegraphics[width=2.5cm,height=1.4cm]{images/gt/umbrella/00019.png} \\

     \includegraphics[width=2.5cm,height=1.4cm]{images/nsff_jumping_media_images_val_rgb_map_blend_40160_9721326c5fdc35459a99.png} & \includegraphics[width=2.5cm,height=1.4cm]{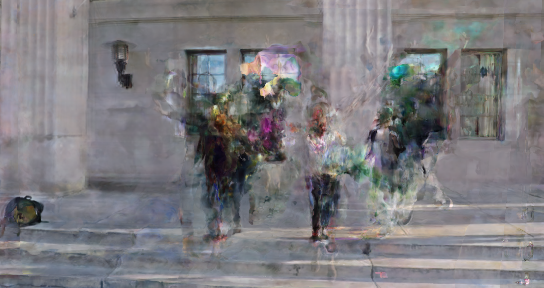} & \includegraphics[width=2.5cm,height=1.4cm]{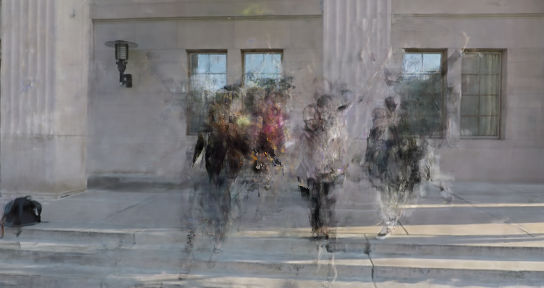} & \includegraphics[width=2.5cm,height=1.4cm]{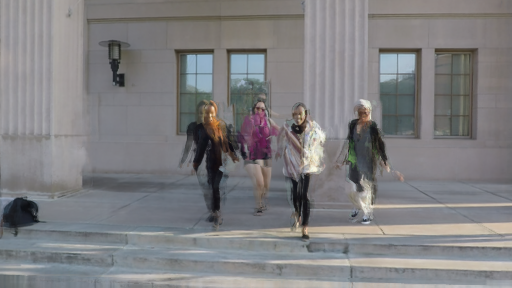} & \includegraphics[width=2.5cm,height=1.4cm]{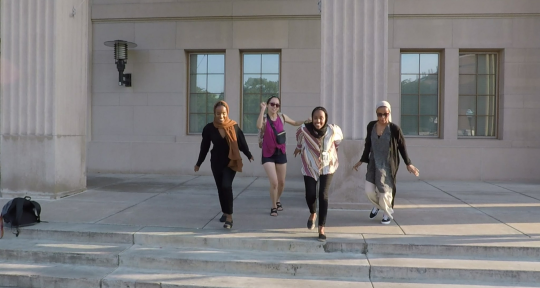}\\

     \includegraphics[width=2.5cm,height=1.4cm]{images/nsff_playground_media_images_val_rgb_map_blend_37716_9ef70b660d1a48430a29.png} & \includegraphics[width=2.5cm,height=1.4cm]{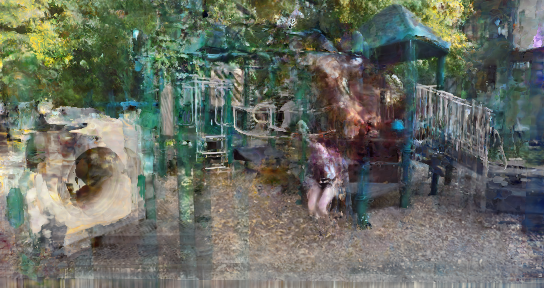} & \includegraphics[width=2.5cm,height=1.4cm]{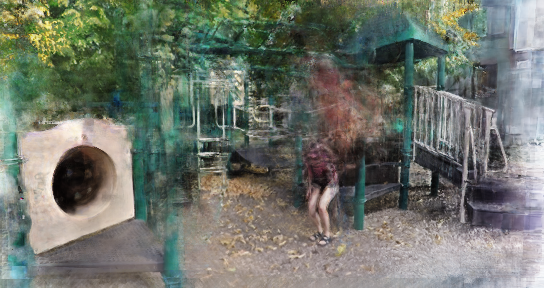} & \includegraphics[width=2.5cm,height=1.4cm]{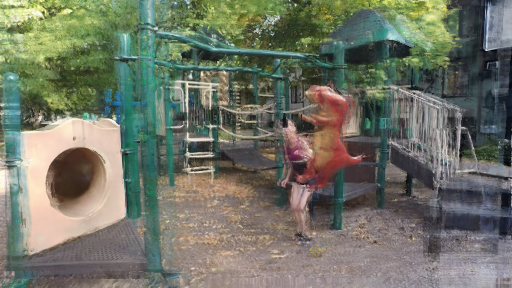} &  \includegraphics[width=2.5cm,height=1.4cm]{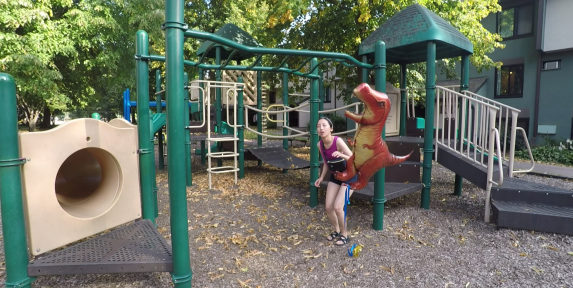}\\

     \includegraphics[width=2.5cm,height=1.4cm]{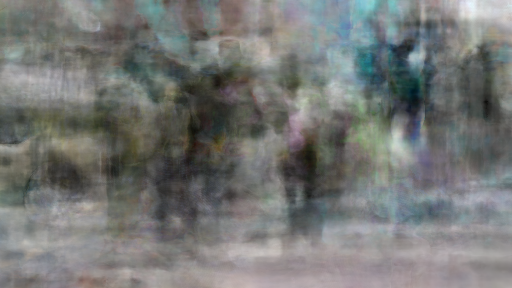} & \includegraphics[width=2.5cm,height=1.4cm]{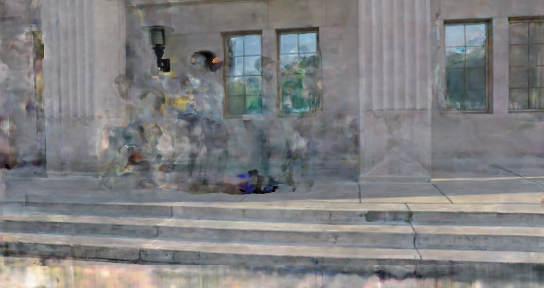} & \includegraphics[width=2.5cm,height=1.4cm]{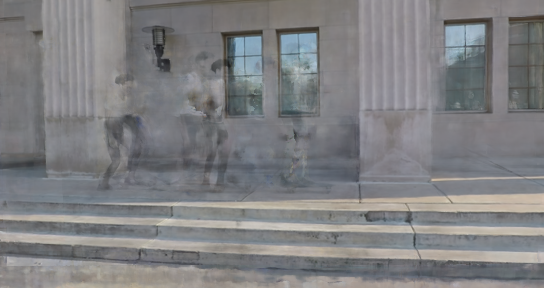} & \includegraphics[width=2.5cm,height=1.4cm]{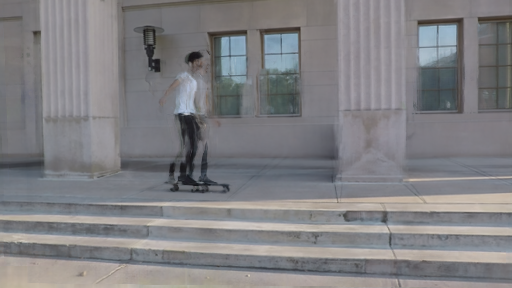} &
     \includegraphics[width=2.5cm,height=1.4cm]{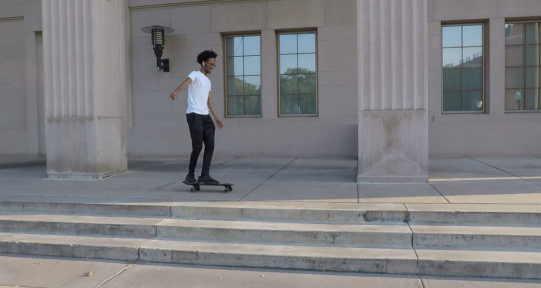}\\

     \includegraphics[width=2.5cm,height=1.4cm]{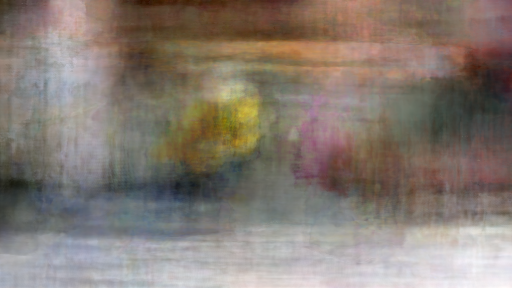} & \includegraphics[width=2.5cm,height=1.4cm]{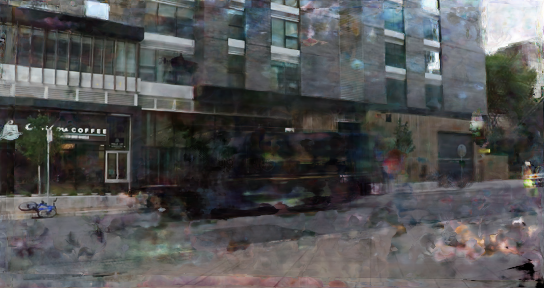} & \includegraphics[width=2.5cm,height=1.4cm]{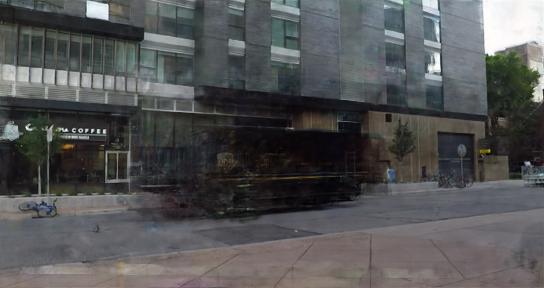} & \includegraphics[width=2.5cm,height=1.4cm]{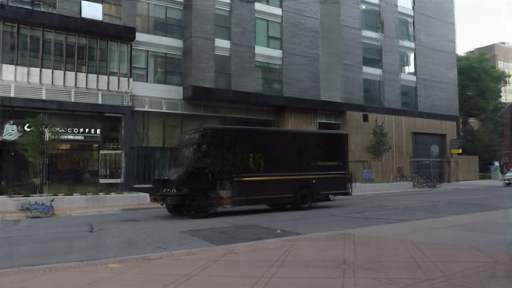} &     \includegraphics[width=2.5cm,height=1.4cm]{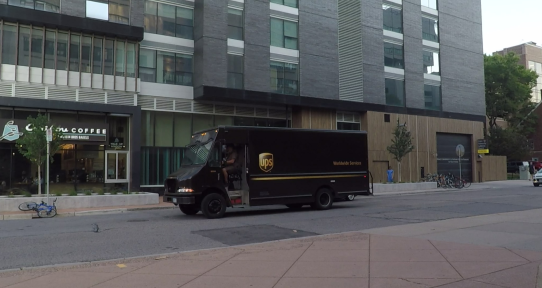} \\
    \end{tabular}
    \vspace{0.2cm}
    \caption{\textbf{Qualitative results} on the Dynamic Scenes dataset~\cite{yoon2020DFNet}}
    \label{fig:qualitative_further}
    % \vspace{-0.5cm}
\end{figure}

For qualitative results, see Figure~\ref{fig:qualitative_further} in Section~\ref{sec:further}.

\newpage

% \vspace{0.5cm}
\section{Further results}
\label{sec:further}

% \end{appendices}

\end{document}